\documentclass[manuscript,nonacm]{acmart}

\AtBeginDocument{%
  \providecommand\BibTeX{{%
    \normalfont B\kern-0.5em{\scshape i\kern-0.25em b}\kern-0.8em\TeX}}}

\setcopyright{acmcopyright}
\copyrightyear{2021}
\acmYear{2021}
\acmMonth{5}
\acmDOI{}

\begin{document}

\title{An Accurate Non-accelerometer-based PPG Motion Artifact Removal Technique using CycleGAN}

\author{Amir Hosein Afandizadeh Zargari}
\authornote{First three authors contributed equally to this research.}
\email{amir.zargari@uci.edu}
\orcid{https://orcid.org/0000-0001-5797-3215}

\author{Seyed Amir Hossein Aqajari}
\authornotemark[1]
\email{saqajari@uci.edu}
\orcid{https://orcid.org/0000-0003-1747-6980}

\author{Hadi Khodabandeh}
\authornotemark[1]
\email{khodabah@uci.edu}
\orcid{https://orcid.org/0000-0003-3850-6739}

\author{Amir M. Rahmani}
\email{a.rahmani@uci.edu}
\orcid{https://orcid.org/0000-0003-0725-1155}

\author{Fadi Kurdahi}
\email{kurdahi@uci.edu}
\orcid{https://orcid.org/0000-0002-6982-365X}

\affiliation{%
  \institution{University of California, Irvine}
  \streetaddress{P.O. Box 92617}
  \city{Irvine}
  \state{CA}
  \country{USA}
  \postcode{43017-6221}
}

\renewcommand{\shortauthors}{}

\begin{abstract}
  A photoplethysmography (PPG) is an uncomplicated and inexpensive optical technique widely used in the healthcare domain to extract valuable health-related information, e.g., heart rate variability, blood pressure, and respiration rate. PPG signals can easily be collected continuously and remotely using portable wearable devices. However, these measuring devices are vulnerable to motion artifacts caused by daily life activities. The most common ways to eliminate motion artifacts use extra accelerometer sensors, which suffer from two limitations: i) high power consumption and ii) the need to integrate an accelerometer sensor in a wearable device (which is not required in certain wearables). This paper proposes a low-power non-accelerometer-based PPG motion artifacts removal method outperforming the accuracy of the existing methods. We use Cycle Generative Adversarial Network to reconstruct clean PPG signals from noisy PPG signals. Our novel machine-learning-based technique achieves 9.5 times improvement in motion artifact removal compared to the state-of-the-art without using extra sensors such as an accelerometer.
\end{abstract}



\maketitle

\section{Introduction}\label{sec:intro}
A photoplethysmography (PPG) is a simple, low-cost, and convenient optical technique used for detecting volumetric blood changes in the microvascular bed of target tissue \cite{allen2007photoplethysmography}. Recently, there has been increasing attention in the literature for extracting valuable health-related information from PPG signals, ranging from heart 
rate and heart rate variability to blood pressure and respiration rate \cite{mehrabadi2020detection}.

Nowadays, PPG signals can easily be collected continuously and remotely using inexpensive, convenient, and portable wearable devices (.e.g., smartwatches, rings, etc.) which makes them a suitable source in wellness applications in everyday life. However, PPG signals collected from portable wearable devices in everyday settings are often measured when a user is engaged with different kinds of activities  and therefore are distorted by motion artifacts. The signal with a low signal-to-noise ratio leads to inaccurate vital signs extraction which may risk life-threatening consequences for healthcare applications. There exists a variety of methods to detect and remove motion artifacts from PPG signals. The majority of the works related to the detection and filtering of motion artifacts in PPG signals can reside in three categories: (1) non-acceleration based, (2) using synthetic reference data, and (3) using acceleration data. 

The non-acceleration based methods do not require any extra accelerometer sensor for motion artifact detection and removal. In existing works, these approaches utilize certain statistical methods due to the fact that some statistical parameters such as skewness and kurtosis will remain unchanged regardless of the presence of the noise. In \cite{cite1}, such statistical parameters are used to detect and remove the impure part of the signal due to motion artifacts. In \cite{cite2}, authors detect motion artifacts using a Variable Frequency Complex Demodulation (VFCDM) method. In this method, the PPG signal is normalized after applying a band-pass filter. Then, to detect motion artifacts, VFCDM distinguishes between the spectral characteristics of noise and clean signals. Then, due to a shift in the frequency, an unclean-marked signal is removed from the entire signal. Another method in this category is proposed in \cite{cite3} that uses the Discrete Wavelet Transform (DWT) method. 

In non-accelerometer based methods, the clean output signal is often shorter than the original signal, since unrecovered noisy data is removed from the signal. To mitigate this problem, a synthetic reference signal can be generated from the corrupted PPG signal. In \cite{cite4}, authors use Complex Empirical Mode Decomposition (CEMD) to generate signals. In \cite{cite5}, two PPG sensors are being used to generate a reference signal. One of the sensors is a few millimeters away from the skin, which only measures PPG during movements. First a band-pass filter is applied on both recorded signals; then, an adaptive filter is used to minimize the difference between two recorded signals. 

Often an accelerometer sensor is also embedded in wearable devices. To eliminate the effect of motion artifacts, acceleration data can be used as a reference signal. In \cite{cite6}, with the help of acceleration data, Singular Value Decomposition (SVD) is used for generating a reference signal for an adaptive filter. Then, the reference signal and PPG signal pass through an adaptive filter to remove motion artifacts. With a similar approach, authors in \cite{cite7} use DC remover using another type of adaptive filter. Another method for motion artifact removal is proposed in \cite{cite8} which follows three steps: (1) signals are windowed, (2) the output signal is filtered, and (3) a Hankel data matrix is constructed. 

Even though using an accelerometer-based method increases the model's accuracy, it suffers from two limitations: i) high power consumption and ii) the need to integrate an accelerometer sensor in a wearable device (which is not required in certain wearables). To overcome these issues, machine learning techniques can be employed as an alternative method to remove noise and reconstruct clean signals \cite{ashrafiamiri2020r2ad,yasaei2020iot}. Furthermore, machine learning techniques are utilized in healthcare domain in processing of a variety of physiological signals such as PPG for data analysis purposes \cite{aqajari2021pain,aqajari2021pyeda,joshi2020machine}. The aim of this paper is to propose a machine learning non-accelemoter-based PPG motion artifacts removal method which is low-power and can outperform the accuracy of the existing methods (even the accelerometer-based techniques).
In recent studies, applying machine learning for image noise reduction has been investigated extensively. The most recent studies use deep generative models to reconstruct or generate clean images \cite{chen2018image,tran2020gan}. In this paper, we propose a novel approach which converts noisy PPG signals to a proper visual representation and uses deep generative models to remove the motion artifacts. We use a Cycle Generative Adversarial Network (CycleGAN) \cite{cite9} to reconstruct clean PPG signals from noisy PPG. CycleGAN is a novel and powerful technique in unsupervised learning, which targets learning the distribution of two given datasets to translate an individual input data from the first domain to a desired output data from the second domain. The advantages of CycleGAN over other existing image translation methods are i) it does not require the pairwise dataset, and ii) the augmentation in CycleGAN makes it practically more suitable for datasets with fewer images. Hence, we use CycleGAN to remove motion artifacts from noisy PPG signals and reconstruct the clean signals. Our experimental results demonstrate the superiority of our approach compared to the state-of-the-art with a 9.5 times improvement with more energy efficiency due to eliminating accelerometer sensors.

The  rest  of  this  paper  is  organized  as  follows. Section Methods introduces the employed dataset and our proposed pipeline architecture. In section Results we summarize the result obtained by our proposed method and compare our result with the state-of-the-art in motion artifact removal from PPG signals. Finally, in the Conclusion section we discuss the strengths and limitations of our method and we cover the future work.

\section{Methods}\label{sec:methods}
\begin{figure}[ht]
    \centering
    \includegraphics[width=0.7\textwidth]{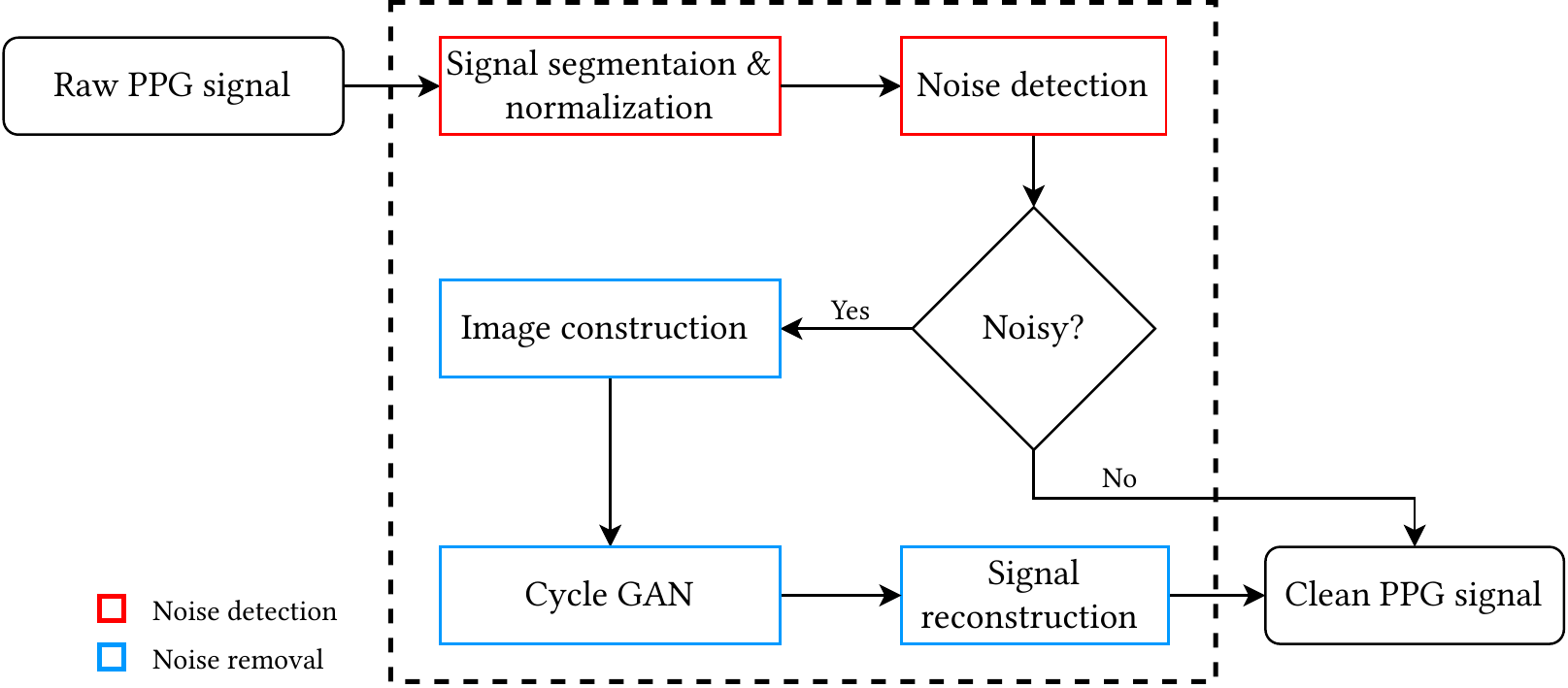}
    \caption{Flowchart of the proposed PPG motion artifacts noise removal}
    \label{fig:1}
\end{figure}

In this paper, we present an accurate non-accelerometer-based motion artifacts removal model from PPG signals. This model mainly consists of a module for noise detection and another one for motion artifact removal. We present in Figure \ref{fig:1} the flow chart of our proposed model. Each module is discussed in detail in the corresponding section.

In order to train this model, two datasets of PPG signals are required: one consisting of clean PPG signals and the other one containing noisy PPG signals. The model's evaluation requires both clean and noisy signals to be taken from the same patient in the same period of time. However, recording such data is not feasible  as patients are either performing an activity, which leads to recording a noisy signal or are in a steady-state, which produces a clean signal. For this reason, we simulate the noisy signal by adding data from an accelerometer to the clean signal. This is a common practice and has been used earlier in related work (e.g., \cite{cite10}) to address this issue. This way, the effectiveness of the model can be evaluated efficiently by comparing the clean signal with the reconstructed output of the model on the derived noisy signal. In the following subsections, we explain the process of data collection for both clean and noisy datasets.

\subsection{BIDMC Dataset}\label{sec:bidmc}
For the clean dataset, we use BIDMC dataset \cite{cite11}. This dataset contains signals and numerics extracted from the much larger MIMIC II matched waveform database, along with manual breath annotations made from two annotators, using the impedance respiratory signal.

The original data was acquired from critically ill patients during hospital care at the Beth Israel Deaconess Medical Centre (Boston, MA, USA). Two annotators manually annotated individual breaths in each recording using the impedance respiratory signal. There are 53 recordings in the dataset, each being 8 minutes long and containing:
\begin{itemize}
    \item Physiological signals, such as the PPG, impedance respiratory signal, and electrocardiogram (ECG) sampled at 125 Hz.
    \item Physiological parameters, such as the heart rate (HR), respiratory rate (RR), and blood oxygen saturation level (SpO2) sampled at 1 Hz.
    \item Fixed parameters, such as age and gender.
    \item Manual annotations of breaths.
\end{itemize}

\subsection{Data Collection}\label{sec:data}
\begin{figure}[ht]
    \centering
    \includegraphics[width=0.8\textwidth]{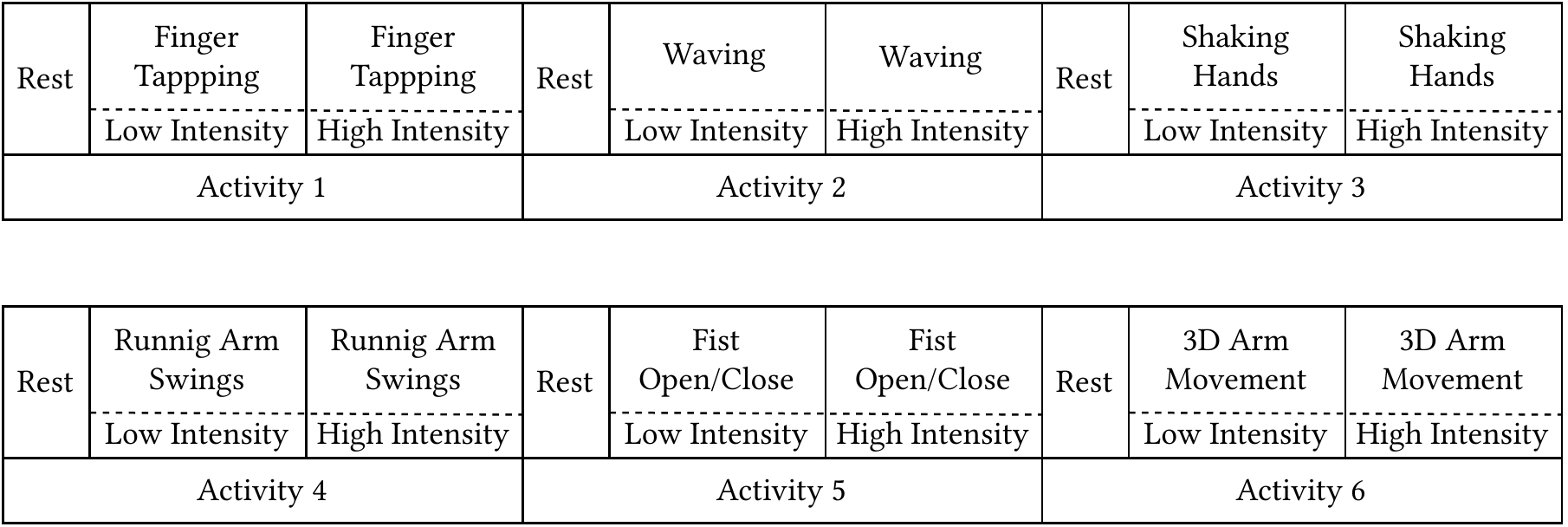}
    \caption{Experimental procedure to collect accelerometer data.}
    \label{fig:2}
\end{figure}

We conducted laboratory-based experiments to collect accelerometer data for generating noisy PPG signals. Each of these laboratory-based experiments consisted of 27 minutes of data. A total of 33 subjects participated in the laboratory-based experiments. In each experiment, subjects were asked to perform specific activities while the accelerometer data were collected from them using an Empatica E4 \cite{cite12} wristband worn on their dominant hand. The Empatica E4 wristband is a medical-grade wearable device that offers real-time physiological data acquisition, enabling researchers to conduct in-depth analysis and visualization. Figure \ref{fig:2} shows our experimental procedure. Note that the accelerometer signals are only required for generating/emulating noisy PPG signals, and our proposed motion artifact removal method does not depend on having access to acceleration signals.

According to Figure \ref{fig:2}, each experiment consists of 6 different activities: (1) Finger Tapping, (2) Waving, (3) Shaking Hands, (4) Running Arm Swing, (5) Fist Opening and Closing, and (6) 3D Arm Movement. Each activity lasts 4 minutes in total, including two parts with two different movement intensities (low and high), each of which lasts 2 minutes. Activity tasks are followed by a 30 seconds rest (R) period between them. During the rest periods, participants were asked to stop the previous activity and put both their arms on a table, and stay in a steady state. Accelerometer data collected during each of the activities were later used to model the motion artifact. We describe this in the next subsection.

\subsection{Noisy PPG signal generation}\label{sec:generation}
To generate noisy PPG signals from clean PPG signals, we use accelerometer data collected in our study. Clean PPG signals are directly collected from the BIDMC dataset. We down-sample these signals to 32 Hz to ensure they are synchronized with the collected accelerometer data.

\begin{figure}[ht]
    \centering
    \includegraphics[trim=0 7cm 0 7cm,clip,width=1\textwidth]{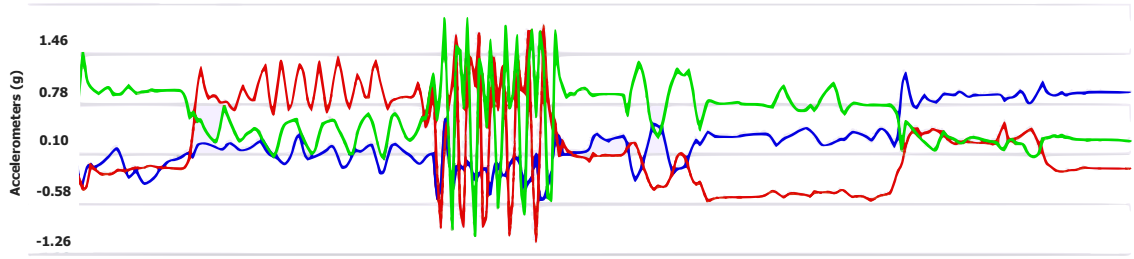}
    \caption{An example of Accelerometer data from Connect, the subject moves into position, walks, runs, and then simulates the turning of a car's steering wheel. The dimensional axes are depicted in red, green and blue.}
    \label{fig:3}
\end{figure}

Empatica has an onboard MEMS type 3-axis accelerometer that measures the continuous gravitational force (g) applied to each of the three spatial dimensions (x, y, and z). The scale is limited to $\pm 2$g. Figure \ref{fig:3} shows an example of accelerometer data collected from E4.

\begin{figure}[ht]
    \centering
    \includegraphics[trim=0 7cm 0 7cm,clip,width=1\textwidth]{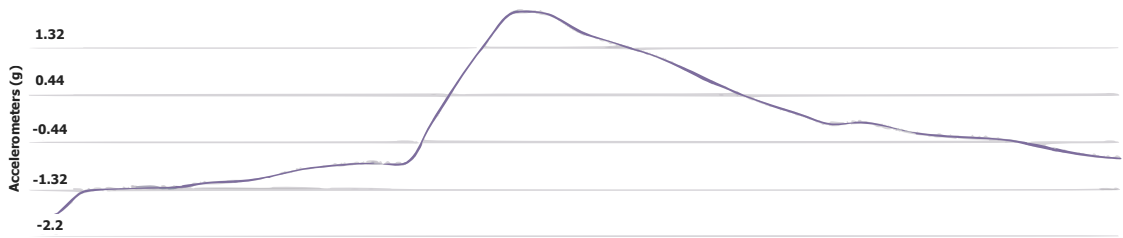}
    \caption{The same data as Figure \ref{fig:3} is visualized using the moving average. From Connect, the subject moves into position, walks, runs, and then simulates the turning of a car's steering wheel. The dimensional axes are depicted in red, green and blue.}
    \label{fig:4}
\end{figure}

Along with the raw 3-dimensional acceleration data, Empatica also provides a moving average of the data. Figure \ref{fig:4} visualizes the moving averaged data.

The following formula is used to calculate the moving averaged of the data,
\begin{equation}
    \text{Sum} += \max(|\text{Curr}_X-\text{Prev}_X|, |\text{Curr}_Y-\text{Prev}_Y|, |\text{Curr}_Z-\text{Prev}_Z|)
    \label{eq:1}
\end{equation}

in which, $\text{Curr}_i$ and $\text{Prev}_i$ are respectively the current value and the previous value of the accelerometer sensor (g) along the $i$-th dimension. The $\max(a,b,c)$ function returns the maximum value among $a$, $b$, and $c$.

Then the following formula is used to filter the output:
\begin{equation}
    \text{Avg} = 0.9 \times \text{Avg}  + 0.1 \times \frac{\text{Sum}}{32} 
    \label{eq:2}
\end{equation}

The filtered output (Avg) is directly used as a model for motion artifacts in our study. To simulate the noisy PPG signals, we add this noise model to a 2 minutes window of the clean PPG signals collected from the BIDMC dataset. We use 40 out of 53 signals in BIDMC directly as the clean dataset for training. Among these 40 signals, 20 are selected and augmented with the accelerometer data to construct the noisy dataset for training. The 13 remaining BIDMC signals and accelerometer data were added together to form the clean and noisy datasets for testing. In the rest of this section we describe each part of the model introduced in Figure \ref{fig:1}.

\subsection{Noise Detection}\label{sec:detection}
To perform noise detection, first, the raw signal is normalized by a linear transformation to map its values to the range $(0,1)$. This can be performed using a simple function as below:
\begin{equation}
    \text{Sig}_{\text{norm}} = \frac{\text{Sig}_{\text{raw}} - \min(\text{Sig}_{\text{raw}})}{\max(\text{Sig}_{\text{raw}}) - \min(\text{Sig}_{\text{raw}})}
    \label{eq:3}
\end{equation}

where $\text{Sig}_{\text{raw}}$ is the raw signal and $\text{Sig}_{\text{norm}}$ is the normalized output. Then, the normalized signal is divided into equal windows of size 256, which is the same window size we use for noise removal. These windows are then used as the input of the noise detection module to identify the noisy ones.

\begin{figure}[ht]
    \centering
    \includegraphics[width=1\textwidth]{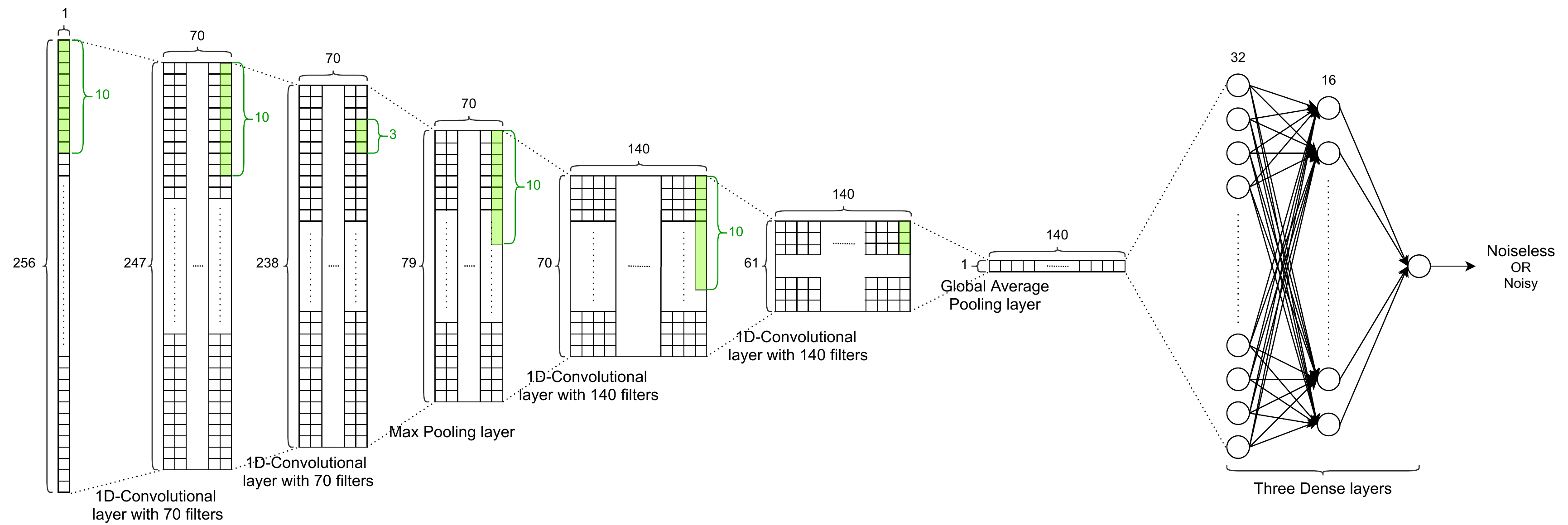}
    \caption{The structure of the noise detection model.}
    \label{fig:5}
\end{figure}

The similar type of machine learning network used in \cite{zargari2020newertrack} can be employed as a noise detection system. To explain the network structure for the noise detection method (Table \ref{tab:layers} and Figure \ref{fig:5}), first, we use a 1D-convolutional layer with 70 initial random filters with a size of 10 to select the basic features of the input data and convert the matrix size from $256\times1$ to $247\times70$. To extract more complex features from the data, another 1D-convolutional layer with the same filter size 10 is required. As the third layer, a pooling layer with a filter size of 3 is utilized. In this layer, a sliding window slides over the input of the layer and in each step, the maximum value of the window is applied to the other values. This layer converts a matrix size of $238\times70$ to $79\times70$. To select additional complex features, another set of convolutional layers are used with a different filter size. This set is followed by two fully connected layers of sizes 32 and 16. Lastly, a dense layer of size 2 with a softmax activation would produce the probability of each class: clean and noisy. The maximum of these two probabilities would be identified as the result of the classification. The accuracy of our proposed binary classification method is 99\%, which means that the system can almost always detect a noisy signal from a clean signal.

\begin{table}[ht]
  \caption{The layer configuration of the noise detection model.}
  \label{tab:layers}
  \begin{tabular}{llc}
    \toprule
    Layer&Structure&Output\\
    \midrule
    Conv1D+Relu&$70\times10$&$247\times70$\\
    Conv1D+Relu&$70\times10$&$238\times70$\\
    Max pooling 1D&3&$79\times70$\\
    Conv1D+Relu&$140\times10$&$70\times140$\\
    Conv1D+Relu&$140\times10$&$61\times140$\\
    Global average pooling&N/A&$140$\\
    Dense+Relu&$128$&$32$\\
    Dense+Relu&$16$&$16$\\
    Dense+Softmax&$2$&$2$\\
  \bottomrule
\end{tabular}
\end{table}

\subsection{Noise Removal}\label{sec:removal}
In this section, we explore the reconstruction of noisy PPG signals using deep generative models. Once a noisy window is detected, it is sent to the noise removal module for further processing. First, the windows are transformed into 2-dimensional images, to exploit the power of existing image noise removal models, and then a trained CycleGAN model is used to remove the noise induced by the motion artifact from these images. In the final step of the noise removal, the image transformation is reversed to obtain the clean output.

The transformation needs to provide visual features for unexpected changes in the signal so that the CycleGAN model would be able to distinguish and hence reconstruct the noisy parts. To extend the 1-dimensional noise on the signal into a 2-dimensional visual noise on the image, we apply the following transformation:
\begin{equation}
    \text{Img}_{i,j} = \operatorname{floor}((\text{Sig}[i]+\text{Sig}[j])\times 128)
    \label{eq:4}
\end{equation}

where Sig is a normalized window of the signal. Each pixel will then have a value between 0 and 255, representing a grayscale image.

\begin{figure}[ht]
    \centering
    \includegraphics[width=0.44\textwidth]{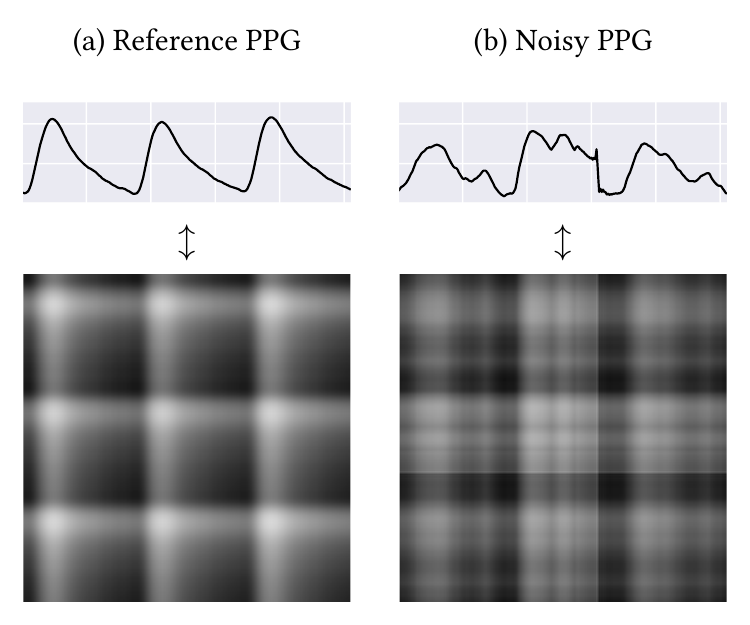}
    \caption{An example of signal to image transformation.}
    \label{fig:6}
\end{figure}

An example of such transformation is provided in Figure \ref{fig:6} for both the clean and the noisy signal. According to this figure, the noise effect is visually observable in these images.

Autoecnoders and CycleGAN are two of the most powerful approaches for image translation. These methods have proven to be effective in the particular case of noise reduction. Autoencoders require the pairwise translation of every image in the dataset. In our case, clean and noisy signals are not captured simultaneously, and their quantity differs. CycleGAN, on the other hand, does not require the dataset to be pairwise. Also, the augmentation in CycleGAN makes it practically more suitable for datasets with fewer images. Hence, we use CycleGAN to remove motion artifacts from noisy PPG signals and reconstruct the clean signals.

CycleGAN is a Generative Adversarial Network designed for the general purpose of image-to-image translation. CycleGAN architecture was first proposed by Zhu et al. in \cite{cite9}.

The GAN architecture consists of two networks: a generator network and a discriminator network. The generator network starts from a latent space as input and attempts to generate new data from the domain. The discriminator network aims to take the generated data as an input and predict whether it is from a dataset (real) or generated (fake). The generator is updated to generate more realistic data to better fool the discriminator, and the discriminator is updated to better detect generated data by the generator network.

The CycleGAN is an extension of the GAN architecture. In the CycleGAN, two generator networks and two discriminator networks are simultaneously trained. The generator network takes data from the first domain as an input and generates data for the second domain as an output. The other generator takes data from the second domain and generates the first domain data. The two discriminator networks are trained to determine how plausible the generated data are. Then the generator models are updated accordingly. This extension itself cannot guarantee that the learned function can translate an individual input into a desirable output. Therefore, the CycleGAN uses a cycle consistency as an additional extension to the model. The idea is that output data by the first generator can be used as input data to the second generator. Cycle consistency is encouraged in the CycleGAN by adding an additional loss to measure the difference between the generated output of the second generator and the original data (and vice versa). This guides the data generation process toward data translation.

In our CycleGAN architecture, we apply adversarial losses \cite{cite13} to both mapping functions ($G: X\to Y$ and $F: Y\to X$). The objective of the mapping function $G$ as a generator and its discriminator $D_Y$ is expressed as below:
\begin{equation}
    L_{GAN}(G, D_Y,X,Y) = E_{y\sim p_{data}(y)}[\log\log D_Y(y)] + E_{x\sim p_{data}(x)}[\log \log (1-D_Y(G(x)))] 
    \label{eq:5}
\end{equation}

where the function $G$ takes an input from domain $X$ (e.g., noisy PPG signals), attempting to generate new data that look similar to data from domain $Y$ (e.g., clean PPG signals). In the meantime, $D_Y$ aims to determine whether its input is from the translated samples $G(x)$ (e.g., reconstructed PPG signals) or the real samples from domain $Y$. A similar adversarial loss is defined for the mapping function $F:Y\to X$ as $L_{GAN}(F, D_X,Y,X)$.

As discussed before, adversarial losses alone cannot guarantee that the learned function can map an individual input from domain X to the desired output from domain $Y$. In \cite{cite9}, the authors argue that to reduce the space of possible mapping functions even further, learned mapping functions ($Y$ and $F$) need to be cycle-consistent. This means that the translation cycle needs to be able to translate back the input from domain $X$ to the original image as $X\to G(X) \to F(G(X)) \sim X$. This is called forward cycle consistency. Similarly, backward cycle consistency is defined as: $y\to F(y)\to G(F(y))\sim y$. This behavior is presented in our objective function as:
\begin{equation}
    L_{\text{cyc}}(G,F)=E_{x\sim p_{data}(x)}[\lVert F(G(x))-x\rVert_1] + E_{y\sim p_{data}(y)}[\lVert G(F(y))-y\rVert_1]
    \label{eq:6}
\end{equation}

Therefore, the final objective of CycleGAN architecture is defined as:
\begin{equation}
    L(G, F,D_X,D_Y)=L_{\text{GAN}}(G, D_Y,X,Y)+L_{\text{GAN}}(F, D_X,Y,X) + \lambda L_{\text{cyc}}(G,F)
    \label{eq:7}
\end{equation}
where $\lambda$ controls the relative importance of the two objectives.

In Equation \ref{eq:7}, $G$ aims to minimize the objective while an adversary $D$ attempts to maximize it. Therefore, our model aims to solve:
\begin{equation}
G^*, F^* = \operatorname{argmin} L(G,F,D_X,D_Y)
\label{eq:8}
\end{equation}

The architecture of the generative networks is adopted from Johnson et al. \cite{cite14}. This network contains four convolutions, several residual blocks \cite{cite15}, and two fractionally-strided convolutions with stride $0.5$. For the discriminator networks, they use $70\times70$ PathGANs \cite{cite16,cite17,cite18}.

After the CycleGAN is applied to the transformed image, the diagonal entries are used to retrieve the reconstructed signal.
\begin{equation}
    \text{Sig}_{\text{rec}}[i]=\text{Img}[i,i]/256
    \label{eq:9}
\end{equation}

\section{Results}\label{sec:results}
In this section, we assess the  efficiency of our model based on the following measures: root mean square error (RMSE) and peak-to-peak error (PPE). A signal window size of 256 and an image size of 256 by 256 were used for all experimental purposes, and 25\% of the data was assigned for validation. The noise detection module had an accuracy of 99\%. The summary of the results for noise removal, including the improvement for each noise type and noise intensity, can be found in Table \ref{tab:results}.

For each noise type, there are two entries in this table, one corresponding to the slow movement and the other one corresponding to the fast movement. The average S/N value for slow movements is $21.7$dB, as provided in the table, while the average S/N value for fast movements is $14.0$dB. For each of the measures, RMSE and PPE, we calculated the error between the generated signal and the reference signal as well as the error between the noisy signal and the reference signal in order to observe the improvement of the model on the noisy signal. The degree of improvement on each noise type is added in a separate column in the table. According to the table, the average of improvement on RMSE is $41\times$ and the average of improvement on PPE is $58\times$.

\def\colen{p{0.09\textwidth}}
\begin{table}[ht]
  \caption{Results of the proposed method.}
  \label{tab:results}
  \begin{tabular}{l\colen\colen\colen\colen\colen\colen\colen}
    \toprule
    Noise Type &	S/N (dB) &	RMSE Gen. (BPM) &	RMSE Nsy. (BPM) &	RMSE$\;\;$ Imprv. &	PPE Gen.$\;\;$ (BPM) &	PPE Nsy.$\;\;$ (BPM) &	PPE Imprv. \\
    \midrule
    Waving &	$20.04$ &	$0.213$ &	$41.76$ &	$196.07\times$ &	$0.136$ &	$32.89$ &	$241.60\times$ \\
    Waving &	$11.30$ &	$2.43$ &	$55.30$ &	$22.75\times$ &	$1.088$ &	$37.90$ &	$34.84\times$ \\
    3D Arm Movement &	$20.17$ &	$1.644$ &	$92.12$ &	$56.03\times$ &	$0.772$ &	$44.03$ &	$57.06\times$ \\
    3D Arm Movement &	$13.12$ &	$1.688$ &	$65.99$ &	$39.10\times$ &	$0.700$ &	$48.49$ &	$69.29\times$ \\
    Shaking Hands &	$21.66$ &	$1.556$ &	$61.89$ &	$39.78\times$ &	$0.576$ &	$28.62$ &	$49.71\times$ \\
    Shaking Hands &	$14.96$ &	$4.203$ &	$84.31$ &	$20.06\times$ &	$2.677$ &	$64.58$ &	$24.12\times$ \\
    Finger Tapping &	$22.99$ &	$1.758$ &	$63.43$ &	$36.07\times$ &	$0.653$ &	$45.14$ &	$69.14\times$ \\
    Finger Tapping &	$13.99$ &	$3.008$ &	$21.76$ &	$7.235\times$ &	$1.191$ &	$10.70$ &	$8.99\times$ \\
    Fist Open Close &	$25.11$ &	$1.648$ &	$35.74$ &	$21.69\times$ &	$0.528$ &	$24.51$ &	$46.44\times$ \\
    Fist Open Close &	$16.69$ &	$2.151$ &	$51.28$ &	$23.84\times$ &	$1.113$ &	$42.65$ &	$38.33\times$ \\
    Running Arm &	$20.14$ &	$2.056$ &	$22.93$ &	$11.16\times$ &	$0.715$ &	$19.32$ &	$27.02\times$ \\
    Running Arm &	$13.98$  &	$3.807$ &	$77.73$ &	$20.42\times$ &	$1.348$ &	$50.75$ &	$37.64\times$\\
    \midrule
    Average &	$17.85$  &	$2.18$ &	$56.19$ &	$41.18\times$ &	$0.958$ &	$37.465$ &	$58.68\times$\\
  \bottomrule
\end{tabular}
\end{table}

\begin{figure}[ht]
    \centering
    \includegraphics[width=0.66\textwidth]{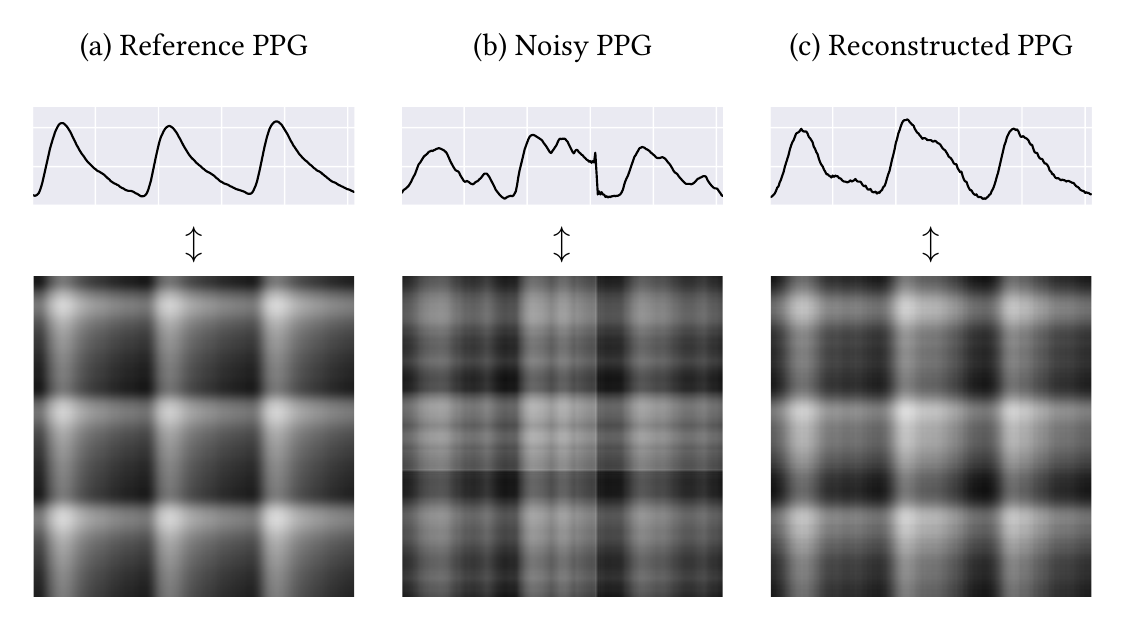}
    \caption{The reconstructed signal of Figure \ref{fig:6} alongside with the noisy and the reference signals}
    \label{fig:7}
\end{figure}

An example of a reconstructed signal is presented in Figure \ref{fig:7}, together with the noisy PPG and the reference PPG signal. As we can see in this figure, the noise is significantly reduced, and the peak values are adjusted accordingly, confirming that the image transformation successfully represents the noise in a visual format.

\subsection{Comparison}\label{sec:comparison}
In this section we compare our model’s efficiency with the state-of-the-art (Table \ref{tab:comparison}). To minimize the difference between our experimental setup and the setups used in the related works we use the same measures. It should be noted that it is not feasible to perform a close comparison between our model and the existing works, due to the differences in the dataset and the lack of a public dataset providing noisy and clean signals simultaneously.

\begin{table}[ht]
  \caption{The summary comparison of our result with the existing methods. MAE stands for Mean absolute error.}
  \label{tab:comparison}
  \begin{tabular}{lllp{0.18\textwidth}p{0.18\textwidth}}
    \toprule
    Paper & Method & Accelerometer & Before & Outcome \\
    \midrule
    \textbf{Proposed method} & \textbf{CycleGAN} & \textbf{No} & \textbf{PPE 37.46 BPM} $\;\;\;$ \textbf{RMSE 56.18 BPM} & \textbf{PPE 0.95 BPM} $\;\;\;$ \textbf{RMSE 2.18 BPM} \\
    Hanyu and Xiaohui \cite{cite1} & Statistical Evaluation & No & PPE 8.1 BPM & PPE 7.85 BPM \\
    Bashar et al. \cite{cite2} & VFCDM & No & N/A & 6.45\% false positive \\
    Lin and Ma \cite{cite3} & DWT & No & PPE 13.97 BPM & PPE 6.87 BPM \\
    Raghuram et al. \cite{cite4} & CEMD LMS  & Syn. & PPE 0.466 BPM & PPE 0.392 BPM \\
    Hara et al. \cite{cite5} & NLMS and RLS & Syn. & RMSE 28.26 BPM & RMSE 6.5 BPM \\
    Tanweer et al. \cite{cite6} & SVD and X-LMS & Yes & N/A & PPE 1.37 BPM \\
    Wu et al. \cite{cite7} & DC remover and RLS & Yes & N/A & STD 3.81 \\
    Bac´a et al. \cite{cite8} & MAR and AT & Yes & N/A & MAE 2.26 BPM \\
  \bottomrule
\end{tabular}
\end{table}

In comparison to non-accelerometer-based methods, our model significantly outperforms these models. The best performance observed in previous work is reported in \cite{cite5} that improves the average RMSE from $28.26$BPM to $6.5$BPM ($4.3\times$ improvement). However, our model's improvement on average RMSE is from $56.18$ to $2.18$ ($25.8\times$ improvement). In most of the existing accelerometer-based methods, no value is provided for the degree of the input noise. Although the best reported PPE belongs to \cite{cite4} with an outcome of $0.392$BPM, the best improvement is achieved by \cite{cite3} from $13.97$BPM to $6.87$BPM ($2.03\times$ improvement). However, our model's improvement on average PPE is from $37.46$BPM to $0.95$BPM ($39.4\times$ improvement).

\section{Discussion}\label{sec:discussion}
Noise reduction has been extensively studied in image processing, and the introduction of powerful models such as CycleGAN has shown promising results in terms of noise reduction in images. Inspired by this fact, we proposed a signal to image transformation that visualizes signal noises in the form of image noise. To the best of our knowledge, this is the first use of CycleGAN for bio-signal noise reduction which eliminates the need for an accelerometer to be embedded into wearable devices, which in turn helps to reduce the power consumption and cost of these devices.

It should be noted that despite the significant benefits of our proposed method in removing noise in different situations, it may not be effective in all possible scenarios. 
Clearly, the intensity of noise applied to the signals, and the variations of the noise, also called noise categories, are controlled for the purpose of this study. When the signal is faded in the noise, this method may not be applicable. Although it will improve the error, it does not guarantee a reasonable upper bound. However, the same limitations also exist in the related works.

\section{Conclusions}
In this paper, we presented an image processing approach to the problem of noise removal from PPG signals where the noise is selected from a set of noise categories that simulate the daily routine of a person. This method does not require an accelerometer on the sensor, therefore, it can be applied to other variations of physiological signals, such as ECG, to reduce the power usage of the measuring device and improve its efficiency. In this work, the novel use of CycleGAN as an image transformer is leveraged to transform such physiological signals. On average, the reconstructed PPG performed using our proposed method offers $41\times$ improvement on RMSE and $58\times$ improvement on PPE, outperforming the state-of-the-art by a factor of $9.5$.

\bibliographystyle{ACM-Reference-Format}
\bibliography{references}
\settopmatter{printacmref=false}










\end{document}